\documentclass[10pt,twocolumn,letterpaper]{article}

 \usepackage[pagenumbers]{cvpr} 

\usepackage[dvipsnames]{xcolor}

\definecolor{cvprblue}{rgb}{0.21,0.49,0.74}
\usepackage[pagebackref,breaklinks,colorlinks,citecolor=cvprblue]{hyperref}

\usepackage[accsupp]{axessibility}
\usepackage[T1]{fontenc}
\usepackage{graphicx}
\usepackage{latexsym}
\usepackage{xspace}
\usepackage{caption}
\usepackage{subcaption}
\usepackage{booktabs} 
\usepackage{multirow}
\usepackage{textcomp}
\usepackage{color, colortbl}
\usepackage[dvipsnames]{xcolor}

\newcommand{\ourapp}{{\sc {Dual-VCR}}\xspace}
\newcommand{\ourappbf}{{\textbf{\textsc {Dual-VCR}}}\xspace}
\newcommand{\mypar}[1]{\vspace{0.5em}\noindent\textbf{#1}.}

\newcommand{\baserank}{{\sc {MindAct\textsubscript{Rank}}}\xspace}
\newcommand{\baserankbf}{{\textbf{\textsc {MindAct\textsubscript{Rank}}}}\xspace}
\newcommand{\basepred}{{\sc {MindAct\textsubscript{Pred}}}\xspace}
\newcommand{\basepredbf}{{\textbf{\textsc {MindAct\textsubscript{Pred}}}}\xspace}

\newcommand{\basepredlarge}{{\sc {MindAct\textsubscript{Pred-large}}}\xspace}

\newcommand{\candonlyvisualrank}{{\sc {Dual-VCR\textsubscript{vis}}}\xspace}
\newcommand{\candonlyvisualrankbf}{{\textbf{\textsc {Dual-VCR\textsubscript{vis}}}}\xspace}

\newcommand{\vneitextrank}{{\sc {Dual-VCR\textsubscript{vnei-txt}}}\xspace}
\newcommand{\vneitextrankbf}{{\textbf{\textsc {Dual-VCR\textsubscript{vnei-txt}}}}\xspace}

\newcommand{\vneitextvisualrank}{{\sc {Dual-VCR\textsubscript{vnei-txt+vis}}}\xspace}
\newcommand{\vneitextvisualrankbf}{{\textbf{\textsc {Dual-VCR\textsubscript{vnei-txt+vis}}}}\xspace}

\newcommand{\vneitextvisualrankweb}{{\sc {Dual-VCR\textsubscript{vnei-txt+vis-web}}}\xspace}
\newcommand{\vneitextvisualrankcoco}{{\sc {Dual-VCR\textsubscript{vnei-txt+vis-coco}}}\xspace}

\newcommand{\vneipred}{{\sc {Dual-VCR\textsubscript{pred}}}\xspace}
\newcommand{\vneipredbf}{{\textbf{\textsc {Dual-VCR\textsubscript{pred}}}}\xspace}

\newcommand{\vneipredlarge}{{\sc {Dual-VCR\textsubscript{pred-large}}}\xspace}

\usepackage{amssymb}
\usepackage{amsmath,amsfonts}
\usepackage{amsopn}
\usepackage{bm} 
\usepackage{multirow}
\usepackage{flushend}
\usepackage{tabularx}

\newcommand{\ProbOpr}[1]{\mathbb{#1}}

\newcommand{\expect}[2]{%
\ifthenelse{\equal{#2}{}}{\ProbOpr{E}_{#1}}
{\ifthenelse{\equal{#1}{}}{\ProbOpr{E}\left[#2\right]}{\ProbOpr{E}_{#1}\left[#2\right]}}} 

\newcommand{\eat}[1]{}

\newcommand{\wholeimagerank}{{\sc {WholeImage\textsubscript{rank}}}\xspace}

\newcommand{\wholevistokrank}{{\sc {WholeVisTok\textsubscript{rank}}}\xspace}

\newcommand{\wholeimagepred}{{\sc {WholeImage\textsubscript{pred}}}\xspace}

\newcommand{\wholehtmlpred}{{\sc {WholeHTML\textsubscript{pred}}}\xspace}

\newcommand{\htmltreeneipred}{{\sc {HTMLTreeNei\textsubscript{pred}}}\xspace}

\newcommand{\randomrank}{{\sc {Random\textsubscript{rank}}}\xspace}

\newcommand{\randompred}{{\sc {Random\textsubscript{pred}}}\xspace}

\def\paperID{17154} 
\def\confName{CVPR}
\def\confYear{2024}

\title{Dual-View Visual Contextualization for Web Navigation}

\author{Jihyung Kil \quad Chan Hee Song \quad Boyuan Zheng \quad Xiang Deng \quad Yu Su \quad Wei-Lun Chao \\ The Ohio State University \\
{\tt\small\{kil.5,song.1855,zheng.2372,deng.595,su.809,chao.209\}@osu.edu}}

\begin{document}
\maketitle
\begin{abstract}
Automatic web navigation aims to build a web agent that can follow language instructions to execute complex and diverse tasks on real-world websites. Existing work primarily takes HTML documents as input, which define the contents and action spaces (\ie, actionable elements and operations) of webpages. Nevertheless, HTML documents may not provide a clear task-related context for each element, making it hard to select the right (sequence of) actions. In this paper, we propose to contextualize HTML elements through their ``dual views'' in webpage screenshots: each HTML element has its corresponding bounding box and visual content in the screenshot. We build upon the insight---\emph{web developers tend to arrange task-related elements nearby on webpages to enhance user experiences}---and propose to contextualize each element with its neighbor elements, using both textual and visual features. The resulting representations of HTML elements are more informative for the agent to take action. We validate our method on the recently released Mind2Web dataset, which features diverse navigation domains and tasks on real-world websites. Our method consistently outperforms the baseline in all the scenarios, including cross-task, cross-website, and cross-domain ones.

\end{abstract}
  
\section{Introduction}
\label{sec:intro}

\begin{figure}
    \centering
    \centerline{\includegraphics[width=\linewidth]{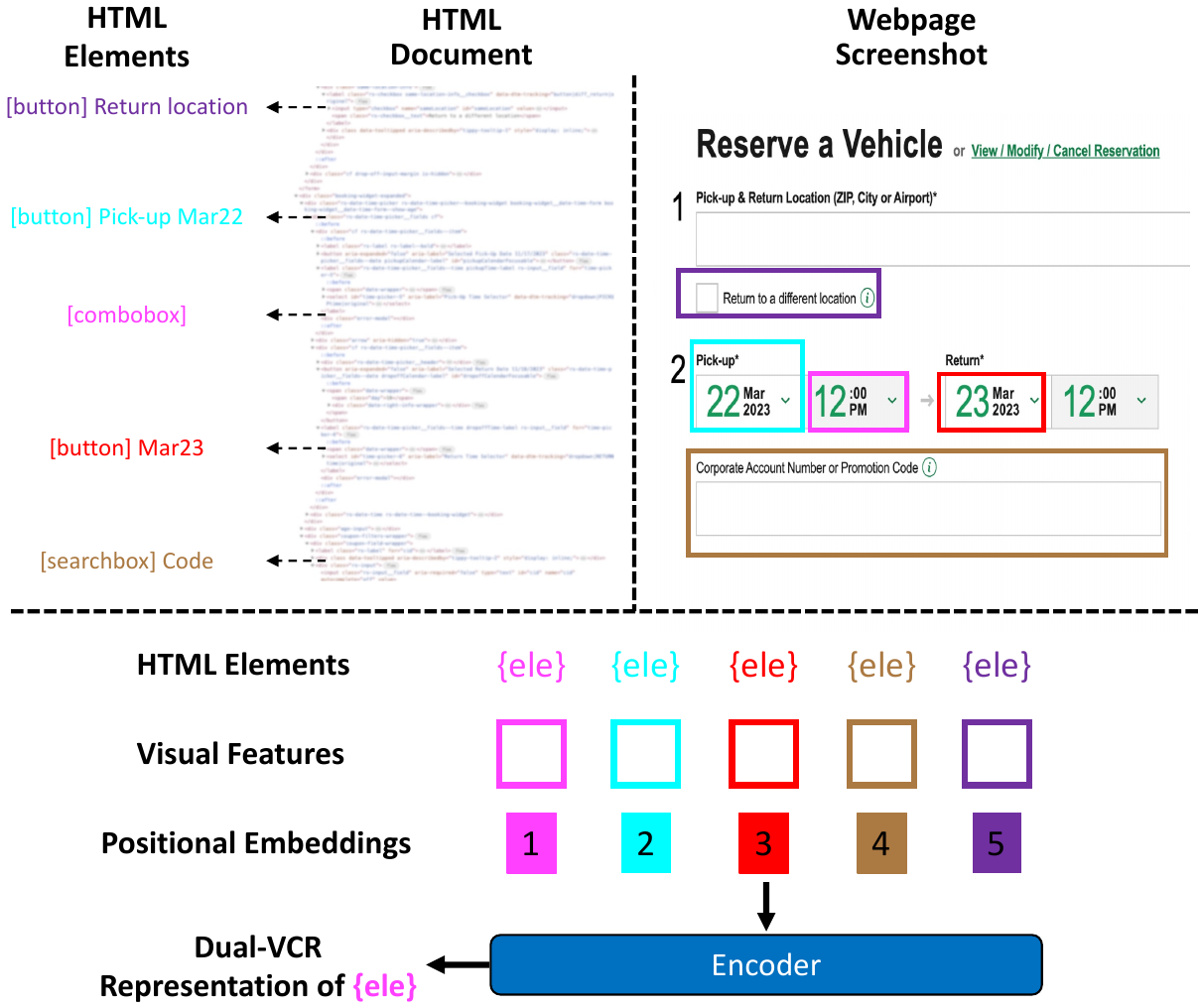}}
    \vspace{-5pt}
    \caption{\textbf{Overview of our proposed Dual-View
Contextualized Representation (\ourappbf).} HTML elements (\eg, {\color{VioletRed}``[combobox]''}) may not have clear contexts for solving web navigation tasks (\eg, ``Find the lowest rent truck with a pick-up time at 11 am on March 27.''). \ourapp contextualizes each element with its neighbors in the screenshot (\eg, {\color{cyan}``[button] Pick-up Mar22''}) to obtain more informative representations for decision-making.}
    \label{fig:overview}
    \vspace{-10pt}
\end{figure}

We study automatic web navigation with natural language instructions~\cite{mind2web,Yao2022WebShopTS}. This problem is crucial as it can potentially streamline and automate a wide range of tasks in our increasingly web-centric world, from online shopping to accessing information. Successfully solving this problem can also broadly advance artificial intelligence as it requires understanding and executing various tasks by interacting with dynamic and complex real-world (web) environments.

Existing work primarily takes HTML documents as the web agent's input~\cite{Gur2023ARW,mind2web,ASH}, which define the meaning and layout of webpage content.
Written partially in natural language, HTML documents enable the use of large language models (LLMs)~\cite{flan-t5,gpt4,gpt3.5,lora,llama2,palm,vicuna2023,alpaca} to ground language instructions (\eg, ``Find one-way flights from New York to Toronto.'')~in web environments. Moreover, elements in HTML documents directly define the space of actions (\eg, element ``[button] Search'' with operation ``click''),
preventing the agent from hallucinating infeasible actions.
 
With that being said, HTML documents may lack a clear task-related context for each element, impeding the agent from 
selecting the right (sequence of) actions to complete a task. HTML is quite flexible for web developers to arrange their code. Even semantically related elements, such as an actionable element (\eg, ``drop-down box'') and its label element (\eg, ``Number of Passengers''), may not be located nearby in the document or the DOM tree. This problem also applies to elements relevant to solving a task.
While LLMs may learn to capture the context, a raw HTML document of real-world webpages is often quite huge, consisting of tens of thousands of tokens, making it
either infeasible or cost-prohibitive to be directly fed into LLMs~\cite{Gur2023ARW,mind2web,ASH}.

In this paper, we propose to enhance the context of each HTML element by leveraging its ``dual view'' in the screenshot of the rendered webpage: many of the HTML elements (including the actionable ones) are visible in the screenshot and have their corresponding bounding boxes\footnote{These bounding boxes can be directly inferred from the HTML document without the need to detect them.}. Taking the insight---\emph{semantically related and task-related HTML elements are often located nearby on the webpage} to facilitate user experiences---we propose to contextualize each HTML element with its neighbors in the screenshot. Concretely, when encoding each HTML element, we 1) append its spatially adjacent elements with positional embeddings and 2) incorporate both the visual and textual features (\autoref{fig:overview}).
 
While simple, our method, which we name \textbf{Dual-View Contextualized Representation (\ourappbf)}, has several compelling properties that benefit web navigation fundamentally. First, \ourappbf uses the built-in feature of HTML documents to align textual and visual content, making it robust to complex and diverse websites. 
Second, \ourappbf effectively leverages visual cues on the webpages, which are designed to ease users' efforts in understanding and completing tasks. 
Specifically, \ourappbf connects \emph{visually proximate elements that are often semantically related and task-related}, providing the agent with more explicit contexts to take not only individual actions but also the sequence of actions. 
Last but not least, \ourappbf can potentially be integrated into any web navigation algorithms that take HTML documents as input.
 
We validate \ourapp on the Mind2Web dataset~\cite{mind2web}, the largest web navigation benchmark with over 2,000 tasks curated from 137 real-world websites across 31 domains, including restaurants, airlines, public services, etc. 
Concretely, we implement \ourappbf on top of the \textbf{MindAct} algorithm \cite{mind2web}, which was proposed to tackle huge HTML documents. In short, at each action, MindAct first applies a small LM to rank each HTML element to shrink the document; it then uses an LLM to predict the action. We integrate \ourapp into both steps to enhance the context for element ranking and decision-making.
\ourapp consistently improves MindAct across all three scenarios (cross-task, cross-website, and cross-domain), leading to a \textbf{3.7}\% absolute gain on average over nine evaluation metrics. 
Moreover, \ourapp notably outperforms baselines that use entire HTML documents or screenshots as input, offering significant advantages in computation and accuracy.

Our contributions are three-folded:
\begin{itemize} [itemsep=2pt,topsep=0.0pt,leftmargin=0pt,partopsep=0pt]
    \item We propose \ourapp, a simple and effective dual-view representation of HTML elements for web navigation.
    \item \ourapp consistently outperforms baselines on the real-world web navigation benchmark Mind2Web~\cite{mind2web}.
    \item We conduct comprehensive analyses to understand the effect of our design choices on web navigation performance.
\end{itemize}

\begin{figure*}
    \centering
    \centerline{\includegraphics[width=0.95\linewidth]{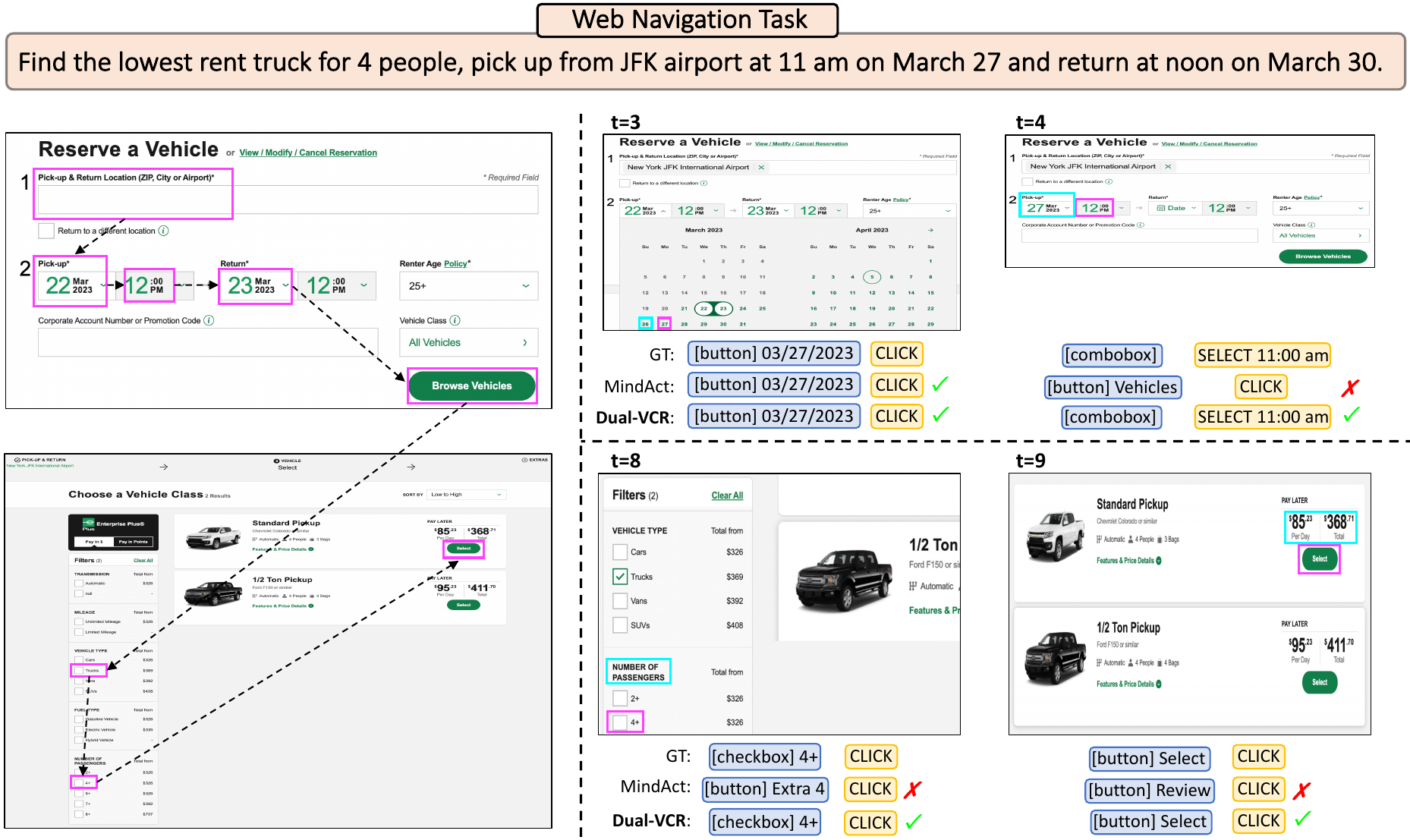}}
    \vspace{-5pt}
    \caption{\textbf{Example of real-world web navigation.} 
    \textbf{Top}: the web navigation task described in natural language.
    \textbf{Left}: the sequence of HTML elements (visualized on webpages, not HTML documents) to interact with to complete the task. We superimpose {\color{magenta}bounding boxes} and arrows to locate the target elements and indicate their order. 
    \textbf{Right}: the detail at each time step (we showed $t=\{3,4,8,9\}$ for brevity). GT: ground-truth action ({\color{blue}Element} with {\color{Goldenrod}Operation}). We compare the predicted actions by MindAct~\cite{mind2web} and our \ourappbf. The {\color{magenta}bounding box} and {\color{cyan}bounding box} indicate the target element and one of its neighbors encoded by \ourappbf. As shown, \ourappbf correctly predicts the elements and operations at ``all'' time steps, taking advantage of the much richer task-related dual-view context it encodes.}
    \vspace{-10pt}
    \label{fig:webnavi_exp}
\end{figure*}

\label{sec:intro}
\section{Related Work}
\label{sec:related}

\mypar{Web navigation datasets}
Several prior studies~\cite{Humphreys2022ADA,Yao2022WebShopTS,pixelhelp,meta-gui,burns2022dataset} have introduced promising benchmarks for assessing agents in web navigation tasks. However, these benchmarks are often limited to a narrow range of website domains or confined to simplified simulated environments. For instance, MiniWob++~\cite{Humphreys2022ADA} and WebShop~\cite{Yao2022WebShopTS} collected a set of websites including daily tasks (\eg, shopping), but each website only has fewer than fifty HTML elements on average. Some other studies~\cite{pixelhelp,meta-gui,burns2022dataset} instead explored other domains, including mobile applications, but their action spaces are often simpler than web navigation. 
Recently, Mind2Web~\cite{mind2web} released the first large-scale web navigation benchmark consisting of over 2K tasks from various real-world websites. This enables a comprehensive understanding of web agent's behaviors in ``real-world'' scenarios.

\mypar{The use of HTML documents}
Most earlier work~\cite{Humphreys2022ADA,Yao2022WebShopTS,liu2018reinforcement,jia2019dom} focused on simple navigation scenarios like MiniWob++~\cite{Humphreys2022ADA}. 
Due to the brevity of its HTML documents, they input whole HTML documents into LLMs to complete the web navigation tasks. A few studies represented HTML documents in a more dense format. For instance, ASH~\cite{ASH} summarized the HTML document using LLMs with hierarchical prompting. DOM-Q-NET~\cite{jia2019dom} leveraged a graph neural network to represent a document as a graph. 
For real-world web navigation (\eg, Mind2Web), HTML documents are often overly lengthy and complex. Thus, recent studies~\cite{mind2web,WebGUM,Gur2023ARW} applied text-based filtering to first identify key HTML elements within the document and only used the selected elements to complete the task.
While all these prior methods are promising, the HTML document alone may not provide a clear task-related context for each element, making it challenging to select the right actions. Our approach instead enhances the context of each HTML element based on their dual view in the screenshot. 

\mypar{The use of webpage screenshots}
Beyond using HTML documents, several studies~\cite{Humphreys2022ADA,Yao2022WebShopTS,pix2act,spotlight,WebGUM,iki2022berts,zheng2024gpt,he2024webvoyager,hong2023cogagent} have explored the incorporation of screenshots for web navigation. Some of them~\cite{Humphreys2022ADA,WebGUM,iki2022berts,he2024webvoyager,hong2023cogagent,zheng2024gpt} utilized both screenshots and HTML documents to learn their joint representations during decision-making. Some others~\cite{pix2act,spotlight,cheng2024seeclick} solely relied on screenshots, bypassing the use of HTML documents. We note that all prior methods primarily focused on utilizing ``whole'' screenshots. 
In contrast, we shift the focus to neighboring elements within the screenshot, providing significant benefits in computation and accuracy.

\section{Approach: \ourappbf}
\label{sec:approach}

We introduce \textbf{Dual-View Contextualized Representation (\ourappbf)} for enhanced web navigation. To begin with, we provide a brief background about web navigation. 

\subsection{Background: web navigation}

A web navigation task consists of a website $S$ (\eg, an airline website) and an instruction $q$ (``Find one-way flights from New York to Toronto.'').
Given $(S, q)$, a web agent $f$ needs to decide and perform a sequence of actions $a=\{a_1, a_2, \cdots, a_t, \cdots\}$ on the website to complete the task. \autoref{fig:webnavi_exp} (left) gives an illustration.

At time step $t$, the website has an HTML document $H_t$, composed of a list of elements $H_t=\{e_{t,1},e_{t,2}, \cdots, e_{t,N}\}$. These HTML elements jointly define 1) the layout and content on the rendered webpage $I_t$, and 2) the action space at time $t$: each candidate action is a pair of an actionable element (\eg, {``[textbox] To''}) and an operation (\eg, ``Type Toronto''). After taking action $a_t$, both the HTML document and webpage will be updated into $(H_{t+1}, I_{t+1})$. For example, clicking the ``[checkbox] One way'' on the airline webpage removes the ``[textbox] Return date'' from the webpage.
Namely, the web environment is dynamic, and the agent must take this into account to decide its actions.
 
Because of the rich content in the HTML document $H_t$, existing work primarily takes it, together with the instruction $q$ and the action history (\eg, \emph{Type New York in the From box}), as the agent's input at time $t$ to decide the next action (\eg, \emph{Type Toronto in the To box}),
\begin{align}
    a_{t+1} = f(q, H_t, \{a_1, a_2, \cdots, a_t\}).
\end{align}
One excellent candidate for $f$ is LLMs~\cite{flan-t5,gpt4,gpt3.5,lora,llama2,palm,vicuna2023,alpaca}, which have shown straggering sucesses in question answering~\cite{wang2022benchmarking} and logical reasoning~\cite{cobbe2021training}. For example, \cite{Humphreys2022ADA,rci} applied LLMs to simplified web navigation. 

\begin{figure}
    \centering
    \centerline{\includegraphics[width=0.95\linewidth]{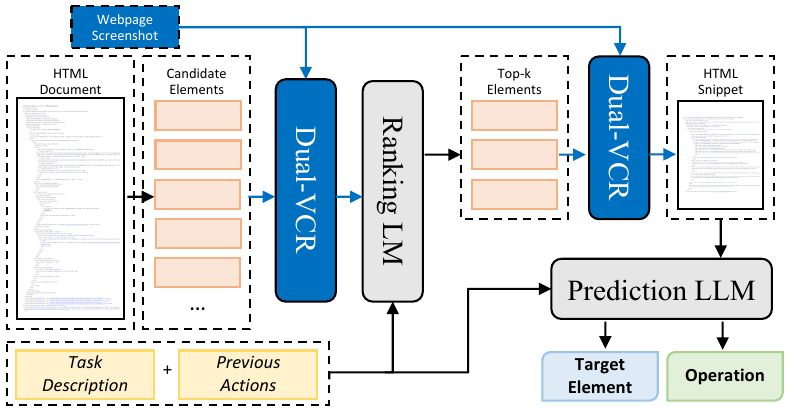}}
    \vspace{-8pt}
    \caption{\textbf{The web navigation pipeline with \ourappbf,} built on top of the MindAct algorithm~\cite{mind2web}. MindAct uses a small ranking LM to select candidate HTML elements and a prediction LLM to decide actions. Blocks and arrows in {\color{NavyBlue} NavyBlue} indicate the insertion of \ourappbf for enhanced element representations.}
    \label{fig:arch}
    \vspace{-12pt}
\end{figure}

However, for real-world webpages that easily contain thousands of HTML elements (amounting to tens of thousands of tokens), directly applying LLMs is neither efficient nor effective. As such, recent work~\cite{Gur2023ARW,mind2web,ASH} employed a two-stage framework: first summarizing the HTML document and then predicting the action. For instance, given the instruction $q$ and the action history at time $t$, the MindAct algorithm~\cite{mind2web} first ranks each HTML element using a small LM. Only the top-$K$ HTML elements are fed into an LLM to predict the next action. (See~\autoref{fig:arch} for an illustration.)

\subsection{Context enhancement}
\label{ssec:context_enh}
We identify one critical pitfall in the two-stage framework. \emph{Since HTML documents may not provide a clear context for each element, the element ranker and the subsequent action predictor may not perform as effectively as expected.}
\autoref{fig:overview} illustrates one such issue: the element {\color{VioletRed}``[combobox]''} should be paired with {\color{cyan}``[button] Pick-up Mar22''} to fully describe its role, \ie, time for pick-up. However, these two elements are not necessarily nearby in the HTML document.

To resolve this issue, we propose to leverage the ``dual view'' of each HTML element $e_{t,n}\in H_t$ in the rendered webpage $I_t$ to enhance its context. In essence, many HTML elements (including the actionable ones) are visible in $I_t$. Further, their visual location (\eg, bounding boxes) can be inferred from HTML documents. Since a webpage (specifically, its screenshot) is designed for users to interact with the website visually, we hypothesize that incorporating the visual cues into HTML element representations would benefit the web agent in understanding and completing tasks.

To this end, we propose \textbf{Dual-View Contextualized Representation (\ourappbf)}. In the screenshot view, we identify the bounding box of each HTML element using a web automation testing tool\footnote{\url{https://playwright.dev/}}.
Taking the insight---web developers tend to arrange semantically relevant and task-related elements in proximity to each other on the screenshot to enhance user experiences---we contextualize each element with its ``visual'' neighbors.
Concretely, we calculate the center points of all elements using their bounding boxes and measure their pairwise distances. For each \emph{candidate} element to be ranked by MindAct, we search for the closest $M$ elements to form its context jointly. 

We consider both the visual and textual information to encode the candidate element and its visual neighbors. We extract each element's visual feature using the Pix2Struct Vision Transformer (ViT)~\cite{pix2struct}, which is pre-trained on webpage screenshots. Specifically, we input the whole screenshot $I_t$ into the ViT and apply ROI Align~\cite{he2017mask,li2022exploring} on top of the output embeddings to obtain the feature vector corresponding to each element's bounding box.
In the HTML document view, we extract each element's corresponding ``HTML text'' following MindAct~\cite{mind2web}. 

\begin{figure}
    \centering
    \centerline{\includegraphics[width=0.95\linewidth]{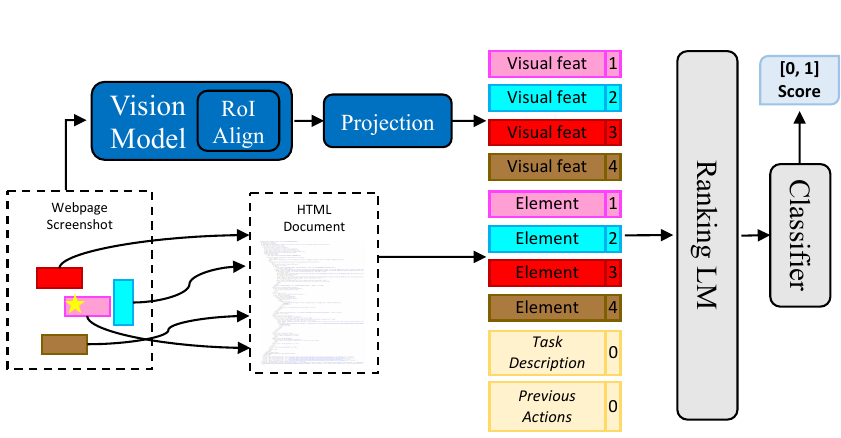}}
    \vspace{-8pt}
    \caption{\textbf{\ourappbf-enhanced element ranker.}. We contextualize the {\color{magenta}candidate} element (denoted by {\color{black}$\star$}) with its neighbors in the screenshot, using both the visual features (by \cite{pix2struct}) and textual features (extracted from the HTML document). Positional embeddings are added to specify neighbor elements, learning their spatial relationships and pairing the textual features with visual features. This dual-view contextualized representation is used to rank the candidate element, measuring its relevance to the current task.
    }
    \label{fig:ranker}
    \vspace{-10pt}
\end{figure}

\subsection{\ourappbf-enhanced element ranker}
\label{ssec:multi_view_ranker}
In MindAct, a small ranking LM is built to predict each element's importance for action prediction. At each time step, the ranking LM takes the element's HTML text tokens, the task description $q$, and the previous actions as input.

We propose to expand the ranking LM to integrate 1) both visual features and textual features and 2) both the candidate element and its neighbor elements. (See~\autoref{fig:ranker} for an illustration.) We make the following design choices. To align the visual embedding and textual embedding, we follow the recent practice of vision-and-language models (\eg, BLIP-2~\cite{BLIP-2}, LLaVA~\cite{LLaVA}, LLaVA-1.5~\cite{LLaVA-1.5}) to learn a linear projection layer to project ViT visual features into the same dimensionality as the token embeddings in the ranking LM. To pair each of the projected visual vectors with its corresponding text tokens and specify each neighbor element in the context, we add positional encoding. Concretely, we sort the neighbors based on their spatial distances from the candidate element and add a learnable positional embedding (unique for each rank) to the neighbor element's visual and text token embeddings. These positionally encoded visual and text token embeddings (of the candidate and the neighbor elements) are fed into the ranking LM; the projected visual features are prepended to the text embeddings, serving as soft visual prompts. In training, we only learn the linear projection layer, the positional embeddings, and the LM while keeping the ViT frozen. This training scheme has been shown to effectively enhance the alignment between vision and language components and improve the pre-trained LM's adaptability to downstream tasks.
Please see more details in the supplementary materials.

\begin{figure}
    \centering
    \centerline{\includegraphics[width=\linewidth]{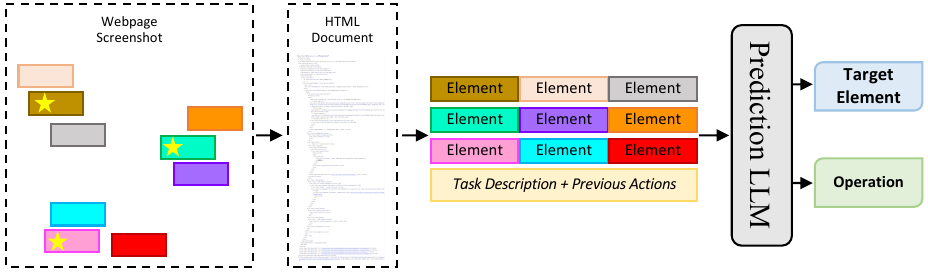}}
    \vspace{-8pt}
    \caption{\textbf{\ourappbf-enhanced action predictor}. Given the top-$K$ candidate elements (three in the figure, marked with {\color{black}$\star$}), \ourapp appends each with its neighbor elements. The resulting HTML snippet, together with the task description and previous actions, is then fed into an LLM for predicting the next action.}
    \label{fig:action_predictor}
    \vspace{-10pt}
\end{figure}

\subsection{\ourappbf-enhanced action predictor}
\label{ssec:action_predictor}
After obtaining the top-$K$ elements from the ranker (\S\ref{ssec:multi_view_ranker}), MindAct combines them into an HTML snippet as the input to LLMs. The objective is to predict the action for the current time step, including the target element (\eg, ``[textbox] To'') and its associated operation (\eg, ``Type Toronto''). 
Specifically, MindAct converts the target element prediction problem into multiple-choice question-answering.

We apply \ourapp to contextualize each of the answer candidates. Similarly to \S\ref{ssec:multi_view_ranker}, we find the $M$ closest neighbors for each candidate element on the screenshot. We then append the HTML text tokens of these $M$ neighbors to the candidate element; we add specific tokens to separate between elements. \autoref{fig:action_predictor} gives an illustration. Please see the supplementary material for more details.

\subsection{Why \ourappbf?}
\ourapp leverages and encodes visual cues on the webpage, offering valuable contexts for the HTML elements in element ranking and action prediction. 
We show two cases. 

First, as shown in~\autoref{fig:overview}, some HTML elements (\eg, {\color{VioletRed}``[combobox]''}) are quite generic and must be paired with spatially nearby elements (\eg, {\color{cyan}``[button] Pick-up Mar22''}) to specify their meanings (\ie, time for pick-up). Similar examples can be found in~\autoref{fig:webnavi_exp}. At $t=8$, there are two seemingly similar candidates ``[checkbox] 4+'' and ``[button] Extra 4''. Nevertheless, the former is spatially closer to the element ``Number of passengers'', indicating its relatedness to the task ``... truck for 4 people ...'' (see the top of \autoref{fig:webnavi_exp}). At $t=9$, two identical ``[button] Select'' elements exist. The only way to differentiate them is through their visual neighbors: one is associated with a lower price than the other. Our \ourapp offers an explicit way to enforce these spatial contexts in the screenshots.

Second, as shown in the left panel of~\autoref{fig:webnavi_exp}, consecutive steps to solve a task often involve spatially nearby elements. Completing one step thus introduces a prior that its nearby elements may be the next to take action upon. As both the ranking LM and prediction LLM take the task description $q$, \emph{past actions}, and our \ourapp representation as input, the models could potentially capture such prior information to increase the success rate for the following action. For example, at $t=4$, \ourapp successfully takes the action ``Select 11:30 am'', likely attributing to its capability to recognize that the previously completed task was the spatially nearby ``Select 03/27/2023''.

\section{Experimental Results}
\label{sec:exp}

\mypar{Dataset}
We validate \ourapp on Mind2Web~\cite{mind2web}, a comprehensive benchmark for real-world web navigation. Unlike other benchmarks based on simulated websites with only a few HTML elements, Mind2Web uses over 100 real-world websites with thousands of HTML elements. Concretely, they provide over 2K open-ended tasks collected from 137 real-world websites across 31 different domains, including travel, shopping, public service, etc (\autoref{tab:stat_mind2web}). Please see more details in the supplementary material.

\mypar{Evaluation Tasks}
Followed by Mind2Web~\cite{mind2web}, we evaluate models at three different test splits. In \textbf{Cross-Domain}, we evaluate the model's generalizability to a new domain where it has not seen any websites or tasks associated with that domain during training. This split contains 912 tasks in total. In \textbf{Cross-Website} (177 tasks), while the model is not exposed to test websites, it is trained on websites from the same domain and potentially with similar tasks. This configuration enables us to evaluate the model's capacity to adapt to entirely new websites within familiar domains and tasks. Similar to the conventional training/test split, \textbf{Cross-Task} (252 tasks) randomly splits 20\% of the data as a test set, regardless of the domains and the websites. Please see the supplementary material for more details.

\mypar{Evaluation Metrics}
We use the Mind2Web's official metrics. The ranker performance is measured by \textbf{Recall@$K$}, where $K$ is the number of top HTML candidate elements.
\textbf{Element Accuracy} (Ele.~Acc) compares the selected element with the ground-truth elements. \textbf{Operation F1} (Op.~F1) calculates the token-level F1 score for the predicted operation. \textbf{Step Success Rate} (Step~SR) measures the success of each step; A step is considered successful only if both the selected element and the predicted operation are correct.  
For each step, they provide previous ``ground-truth'' actions with the assumption that the model successfully completes all previous steps.  
\label{sec:exp}

\mypar{Baselines}
\label{sec:baselines}
\ourapp is based on MindAct~\cite{mind2web}, which has a ranking LM and a prediction LLM. Our main baselines are thus its ranker and action predictor, denoted by \baserankbf and \basepredbf. \baserank uses DeBERTa\textsubscript{base}~\cite{he2020deberta}, a small encoder-only LM to rank elements. For action prediction, \basepred uses Flan-T5\textsubscript{base}~\cite{flan-t5}, an instruction fine-tuned LLM. 


\mypar{Our Models}
\label{sec:models}
Aligned with MindAct, we use the same DeBERTa\textsubscript{base}~\cite{he2020deberta} / Flan-T5\textsubscript{base}~\cite{flan-t5} for our ranker / action predictor, repsectively. For visual features extraction, we utilize Pix2Struct~\cite{pix2struct}'s ViT (pre-trained on screenshots) as the visual backbone and apply ROI Align~\cite{he2017mask} on the element's region. 
We use two linear layers to project visual features into textual embedding space. 
Please see the supplementary materials for details on the model training.

\mypar{Notation of \ourappbf}
\label{sec:our_model_notation}
\ourapp has several variations to understand the effect of each of its components in detail. We denote them as follows:
\begin{itemize} [itemsep=3pt,topsep=0.0pt,leftmargin=0pt,partopsep=0pt]
    \item \small \textbf{\candonlyvisualrankbf}: Ranker w/ candidate's visual features.
    \item \small \textbf{\vneitextrankbf}: Ranker w/ neighbors' HTML text.
    \item \small \textbf{\vneitextvisualrankbf}: Ranker w/ candidate's visual features and its neighbors' visual features and HTML text.
    \item \small \textbf{\vneipredbf}: Action predictor w/ neighbors' HTML text.
\end{itemize}


\subsection{Effectinvess of \ourappbf}
\label{ssec:exp_res}

The main goal of our experiments is to show that our dual-view contexutalization is beneficial in (i) finding promising top-$K$ candidates from entire HTML documents (\ie, ranking peformance), and (ii) predicting the action, including both element selection and operation prediction. 

\begin{table}[t]
\centering
\small
\tabcolsep 1.0pt
\renewcommand\arraystretch{1.0}
\begin{tabular}{lccccc}
\toprule
\multirow{2}{*}{Dataset} & \multirow{2}{*}{\# Websites} &\multirow{2}{*}{Website} & \multirow{2}{*}{\# Tasks} & \multicolumn{2}{c}{Avg \# HTML} \\
\cmidrule(r){5-6}
& & Type & & Elements & Tokens \\
\midrule
MiniWoB++~\cite{Humphreys2022ADA} & 100 & Simplified & 100 & 28 & 500 \\
Mind2Web~\cite{mind2web} & 137 & Real-world & 2,350 & 1,135 & 44,402 \\
\bottomrule
\end{tabular}
\vspace{-5pt} 
\caption{\small \textbf{Statistics of Mind2Web~\cite{mind2web}.} Min2Web, the largest web navigation benchmark, collects real-world websites across various domains. The significant volume of content on the webpage (\eg, an average of 1K/44K HTML elements/tokens) poses challenges for LLMs in both computational and learning aspects.
}
\vspace{-5pt}
\label{tab:stat_mind2web}
\end{table}

\begin{table}[t]
\centering
\small
\tabcolsep 8.5pt
\renewcommand\arraystretch{1.0}
\begin{tabular}{lcccc}
\toprule
\multirow{2}{*}{Ranker} & \multicolumn{4}{c}{Recall} \\
\cmidrule{2-5}
& @1 & @5 & @10 & @50 \\
\midrule
\baserank & 25.4 & 61.0 & 73.5 & 88.9 \\
\vneitextrank & 37.3 & 70.8 & 79.3 & 89.2 \\
\candonlyvisualrank & 37.1 & 70.2 & 79.2 & 89.1 \\
\vneitextvisualrank & \textbf{38.4} & \textbf{71.6} & \textbf{79.7} & \textbf{90.1} \\
\bottomrule
\end{tabular}
\vspace{-5pt} 
\caption{\small \textbf{Ranking performance.} Visual neighbors' HTML text (\vneitextrank) consistently outperforms \baserank. Moreover, \vneitextvisualrank, using both visual neighbors' HTML text and visual features, performs best, showing the strength of dual-view contextualization in element ranking.}
\vspace{-10pt}
\label{tab:rank}
\end{table}

\begin{table*}[t]
\centering
\small
\tabcolsep 3.5pt
\renewcommand\arraystretch{1.0}
\begin{tabular}{llccccccccc}
\toprule
\multirow{2}{*}{Ranker} &
\multirow{2}{*}{Action} &
\multicolumn{3}{c}{Cross-Task} &
\multicolumn{3}{c}{Cross-Website} &
\multicolumn{3}{c}{Cross-Domain} \\
\cmidrule(r){3-5} \cmidrule(r){6-8} \cmidrule(r){9-11}
& Predictor & Ele. Acc & Op. F1 & Step SR & Ele. Acc & Op. F1 & Step SR & Ele. Acc & Op. F1 & Step SR \\
\midrule
\baserank & \basepred & 42.0 & 74.9 & 41.1 & 30.7 & 67.0 & 30.0 & 31.5 & 66.6 & 31.0 \\
\midrule
\vneitextrank & \multirow{2}{*}{\vneipred} & 45.3 & 78.4 & 44.5 & 32.0 & 71.5 & 31.5 & 32.4 & 72.9 & 32.0 \\
\vneitextvisualrank & & \textbf{47.0} & \textbf{78.7} & \textbf{46.0} & \textbf{32.7} & \textbf{72.0} & \textbf{32.5} & \textbf{33.2} & \textbf{73.3} & \textbf{32.5} \\
\bottomrule
\end{tabular}
\vspace{-5pt} 
\caption{\small \textbf{Results of action prediction.} Our \vneitextrank\textrightarrow~\vneipred, leveraging visual neighbors' HTML text information, notably improves over the baseline (\baserank\textrightarrow\basepred) on all nine metrics. Adding visual neighbors' visual features (\vneitextvisualrank) leads to further improvements, highlighting the benefit of dual-view context on real-world web navigation.}
\vspace{-10pt}
\label{tab:main}
\end{table*}

\mypar{Ranking performance}
\autoref{tab:rank} summarizes the ranking results across different top-$K$ candidate elements. First, we see that incorporating the visual neighbor elements' HTML text (\vneitextrank) consistently and significantly outperforms \baserank on all Recall@$K$s (\eg, 37.3\% vs.~25.4\% on Recall@1, 79.3\% vs.~73.5\% on Recall@10), suggesting that contextualizing the element with its neighbors indeed helps find the target element. Second, the candidate element's visual features (\candonlyvisualrank) lead to notable improvements over \baserank (\eg, 70.2\% vs.~61.0\% on Recall@5). This implies that the visual features offer additional context in differentiating HTML elements, compared to using only its HTML text. Lastly, \vneitextvisualrank achieves a further boost by leveraging both visual neighbors' HTML text and visual features (\eg, 38.4\%/90.1\% on Recall@1/@50).

\mypar{Action prediction performance}
\autoref{tab:main} shows the results of action prediction. Compared to the baseline (the combination of~\baserank and~\basepred), using the visual neighbors' HTML texts (\vneitextrank\textrightarrow~\vneipred) notably improves across all metrics. For instance, we achieve gains of 3.4\% on Step~SR in Cross-Task, 1.3\% on Ele.~Acc in Cross-Webiste, and 6.3\% on Op.~F1 in Cross-Domain. These consistent improvements demonstrate the advantages of incorporating visual neighbor information during the model's decision-making process. 
Moreover, aligning with the ranking result, integrating the visual neighbors' visual features into the ranker (\vneitextvisualrank) shows its effectiveness in action prediction as well. Concretely, it achieves the best performance on all nine metrics, along with a 5\% maximum gain on each type of metric against the baseline (\eg, Ele.~Acc: 47.0\% vs.~42.0\% on Cross-Task, Op.~F1: 72.0\% vs.~67.0\% on Cross-Website, Step~SR: 46.0\% vs.~41.1\% on Cross-Task). 

\subsection{Analysis}
\label{ssec:exp_anal}
We aim to understand \ourapp in detail. We show a) a more in-depth analysis of the main table, b) the interaction between the ranker and the action predictor, c) its effectiveness compared to whole input data and random elements, and d) the effect of different sizes of visual neighbors.

\begin{table}[t]
\centering
\small
\tabcolsep 1.0pt
\renewcommand\arraystretch{1.0}
\begin{tabular}{llccc}
\toprule
\multirow{2}{*}{Ranker} &
\multirow{2}{*}{Action} &
\multicolumn{3}{c}{Cross-Task} \\
\cmidrule(r){3-5}
& Predictor & Ele. Acc & Op. F1 & Step SR \\
\midrule
\baserank & \multirow{4}{*}{\basepred} & 42.0 & 74.9 & 41.1 \\
\candonlyvisualrank & & 42.5 & 75.1 & 41.5 \\
\vneitextrank & & 44.6 & 75.7 & 43.2 \\
\vneitextvisualrank & & 46.0 & 78.6 & 44.8 \\
\midrule
\baserank & \multirow{4}{*}{\vneipred} & 44.4 & 75.2 & 43.1 \\
\candonlyvisualrank & & 44.6 & 76.8 & 43.8 \\
\vneitextrank & & 45.3 & 78.4 & 44.5 \\
\vneitextvisualrank & & \textbf{47.0} & \textbf{78.7} & \textbf{46.0} \\
\bottomrule
\end{tabular}
\vspace{-5pt} 
\caption{\small \textbf{Ablation studies} for validating the importance of each component in \ourapp. See \S\ref{ssec:exp_anal} for a detailed discussion.}
\label{tab:main_abl}
\vspace{-10pt}
\end{table}

\mypar{Detailed ablation}
\autoref{tab:main_abl} provides more details about the main table to better understand the impact of each component in \ourapp. First, we keep the action predictor as \basepred and focus on the pure effects of our rankers on the action prediction task (\ie, 1st to 4th rows). We see that incorporating the candidate element's visual features (\candonlyvisualrank) achieves a slight but significant improvement over \baserank across all metrics (\eg, 42.5\% vs.~42.0\% on Ele.~Acc).
Furthermore, our ranker with the visual neighbors' HTML text (\vneitextrank) outperforms \baserank by a notable margin of +2.6\%/+0.8\%/+2.1\% on Ele.~Acc/Op.~F1/Step~SR, respectively. Besides, \vneitextvisualrank, which encodes the visual neighbors' visual features, further improves the model's decision-making ability (\eg, 46.0\% vs.~44.6\% on Ele.~Acc). In short, we consistently demonstrate the effectiveness of each component in our ranker.

Second, conversely, we fix the ranker and examine the benefit of encoding visual neighbors' HTML text features into the action predictor (\vneipred). Compared to \basepred, \vneipred achieves consistent gains across all rankers. For instance, \baserank\textrightarrow~\vneipred outperforms \baserank\textrightarrow~\basepred (\eg, 44.4\% vs.~42.0\% on Ele.~Acc). Similarly, when fixing the ranker with \vneitextvisualrank, \vneipred improves over \basepred (\eg, 46.0\% vs.~44.8\% on Step~SR). This shows directly encoding the visual neighbor's HTML text into the action predictor is beneficial.

Finally, \vneitextvisualrank and \vneipred are complementary; we achieve the best performance across all metrics when leveraging both (\eg, 47.0\%/78.7\%/46.0\% on Ele.~Acc/Op.~F1/Step~SR).
Please see more ablation studies in the supplementary materials.

\begin{table*}[t]
\centering
\small
\tabcolsep 1.5pt
\renewcommand\arraystretch{1.0}
\begin{tabular}{llcccccccccccc}
\toprule
\multirow{2}{*}{Ranker} &
\multirow{2}{*}{Action} &
\multicolumn{3}{c}{Top-1} &
\multicolumn{3}{c}{Top-5} &
\multicolumn{3}{c}{Top-10} &
\multicolumn{3}{c}{Top-50} \\
\cmidrule(r){3-5} \cmidrule(r){6-8} \cmidrule(r){9-11} \cmidrule(r){12-14}
& Predictor & Recall & Ele. Acc & Op. F1 & Recall & Ele. Acc & Op. F1 & Recall & Ele. Acc & Op. F1 & Recall & Ele. Acc & Op. F1 \\
\midrule
\baserank & \multirow{2}{*}{\basepred} & 25.4 & 24.0 & 23.7 & 61.0 & 39.2 & 52.1 & 73.5 & 41.4 & 62.8 & 88.9 & 42.0 & 74.9 \\
\vneitextrank & & \textbf{37.3} & \textbf{35.5} & \textbf{33.5} & \textbf{70.8} & \textbf{43.1} & \textbf{54.1} & \textbf{79.3} & \textbf{43.9} & \textbf{63.0} & \textbf{89.2} & \textbf{44.6} & \textbf{75.7} \\
\bottomrule
\end{tabular}
\vspace{-5pt} 
\caption{\small \textbf{Relationship between ranker and action predictor on Cross-Task.} The ranker has a linear correlation with the action predictor, suggesting the importance of improving its ranking capabilities for decision-making.}
\label{tab:rel_rank_pred}
\vspace{-10pt}
\end{table*}

\mypar{Ranker-action predictor relationship}
We analyze the relationship between the ranker and the action predictor in ~\autoref{tab:rel_rank_pred}. We observe a linear connection between the two. Concertely, improving the ranker (\eg, 25.4\% vs.~37.3\% on Recall@1) correlates with improved action prediction results (\eg, 24.0\% vs.~35.5\% on Ele.~Acc). Aligned with results in \S\ref{ssec:exp_anal}, this again highlights the importance of improving the model's ranking ability in web navigation. 

\mypar{Comparison to whole input data}
Since HTML documents contain a significant amount of content, such as thousands of HTML elements, conducting experiments with whole data is computationally challenging. Nevertheless, we do our best to report the associated results on~\autoref{tab:whole_data} to give more context on the effect of \ourapp. First, instead of asking the ranker to prune HTML documents, we directly pass the whole HTML documents into the action predictor (\wholehtmlpred). We see that \wholehtmlpred performs notably less against the baseline (\basepred) (\ie, 38.6\% vs.~42.0\% on Ele.~Acc). We attribute this to the difficulty of finding the target element among \emph{all thousands} of elements. In contrast, our \vneipred achieves a much better result (\ie, 44.4\%) with significantly less amount of input elements.  

Second, \ourapp outperforms the utilization of whole images. We first use the entire image for the ranker (\wholeimagerank). To extract the image features, we use the same procedure mentioned in~\S\ref{ssec:context_enh}, except for providing the region of the whole image instead of that of specific elements. 
We then use these whole image features, along with the same HTML text input used in \basepred, to train \wholeimagerank. Although the entire image features are shown effective over the baseline (\ie, 43.9\% vs.~42.0\%), it performs notably less than our approach using the \emph{visual neigbhor}'s visual information (\ie, 46.0\% of \vneitextvisualrank). In addition, we conducted a study applying the whole image to the action predictor. Specifically, similar to recent vision-and-language models~\cite{BLIP-2, LLaVA, LLaVA-1.5}, we extract whole image features
using fine-tuned ViT~\cite{pix2struct} and prepend them to the top-50 candidate elements extracted from \baserank as the input to the LLM (Flan-T5\textsubscript{base}~\cite{flan-t5}). 
Similar to the result of \wholeimagerank, this action predictor (\wholeimagepred) performs worse than \vneipred, which only uses \emph{visual neighbors}' HTML text. Overall, this highlights the advantages of our approach in terms of computational efficiency and performance. See additional results in the supplementary materials.

\begin{table}[t]
\centering
\small
\tabcolsep 4pt
\renewcommand\arraystretch{1.0}
\begin{tabular}{llc}
\toprule
\multirow{2}{*}{Ranker} &
Action &
Cross-Task \\
& Predictor & Ele. Acc \\
\midrule
\multirow{3}{*}{\baserank} & \basepred & 42.0 \\
& \wholeimagepred & 43.6 \\
& \vneipred & 44.4 \\
\midrule
\wholeimagerank & \multirow{3}{*}{\basepred} & 43.9 \\
\vneitextrank & & 44.6 \\
\vneitextvisualrank & & \textbf{46.0} \\
\midrule
\quad - & \wholehtmlpred & 38.6 \\
\bottomrule
\end{tabular}
\vspace{-5pt} 
\caption{\small \textbf{Visual neighbor vs.~whole input data.} Using visual neighbors notably outperforms the use of whole data, offering advantages regarding computational efficiency and performance.}
\label{tab:whole_data}
\vspace{-5pt}
\end{table}

\begin{table}[t]
\centering
\small
\tabcolsep 2pt
\renewcommand\arraystretch{1.0}
\begin{tabular}{lclcc}
\toprule
\multirow{2}{*}{Ranker} &
\multirow{2}{*}{Recall} &
\multirow{2}{*}{Action} &
\multicolumn{2}{c}{Cross-Task} \\
\cmidrule(r){4-5}
& @50 & Predictor & Ele. Acc & Op. F1 \\
\midrule
\multirow{3}{*}{\baserank} & \multirow{3}{*}{88.9} & \basepred & 42.0 & 74.9 \\
& & \randompred & 41.5 & 73.6 \\
& & \vneipred & 44.4 & 75.2 \\
\midrule
\randomrank & 86.7 & \multirow{2}{*}{\basepred} & 40.6 & 72.0 \\
\vneitextrank & \textbf{89.2} & & \textbf{44.6} & \textbf{75.7} \\
\bottomrule
\end{tabular}
\vspace{-5pt} 
\caption{\small \textbf{Visual neighbors vs.~random elements.} Visual neighbors provide meaningful contexts for web navigation, notably outperforming elements randomly extracted from HTML documents.}
\vspace{-5pt}
\label{tab:rand_ele}
\end{table}

\mypar{Visual neighbors offer meaningful contexts}
We examine whether visual neighbors provide meaningful context for element ranking and action prediction. To assess this, we compare visual neighboring elements with random elements (\autoref{tab:rand_ele}). Specifically, We randomly select (five) elements from HTML documents and use them to train either the ranker or the action predictor. While our ranker (\eg, \vneitextrank) notably improves the ranking performance over \baserank (\eg, 89.2\% vs.~88.9\%), the ``random'' ranker performs less than \baserank (\eg, 86.7\% vs.~88.9\%). This, in turn, leads to a significant performance drop in the action prediction (\eg, 42.0\% vs.~40.6\% on Ele.~Acc). Similarly, compared to the \basepred, including random elements in the action predictor hurts the action prediction performance (\eg, 74.9\% vs.~73.6 on Op.~F1) while visual neighbors are beneficial (\eg, 75.2\%). In sum, we empirically demonstrate the benefits of context in visual neighbors for web navigation.

\begin{table}[t]
\centering
\small
\tabcolsep 1.5pt
\renewcommand\arraystretch{1.0}
\begin{tabular}{lcccc}
\toprule
\multicolumn{3}{c}{Ranker} &
\multicolumn{2}{c}{Cross-Task} \\
\cmidrule(r){1-3} \cmidrule(r){4-5}
Method & \# neighbors & Recall@50 & Ele. Acc & Op. F1 \\
\midrule
\candonlyvisualrank & 0 & 89.1 & 42.5 & 75.1 \\
\midrule
\multirow{3}{*}{\vneitextvisualrank} & 3 & 89.7 & 45.5 & 77.3 \\
& 5 & \textbf{90.1} & \textbf{46.0} & \textbf{78.6} \\
& 10 & 89.5 & 45.2 & 77.0 \\
\bottomrule
\end{tabular}
\vspace{-5pt} 
\caption{\small \textbf{Effects of the number of neighbors on ranker.} Choosing the right size of visual neighbors is important for element ranking, and the size of five is found to be most effective for Mind2Web~\cite{mind2web}. We fix the action predictor with \basepred.}
\label{tab:num_nei_rank}
\vspace{-15pt}
\end{table}

\begin{table}[t]
\centering
\small
\tabcolsep 6.5pt
\renewcommand\arraystretch{1.0}
\begin{tabular}{lccc}
\toprule
\multicolumn{2}{c}{Action Predictor} &
\multicolumn{2}{c}{Cross-Task} \\
\cmidrule(r){1-2} \cmidrule(r){3-4}
Method & \# neighbors & Ele. Acc & Op. F1 \\
\midrule
\basepred & 0 & 46.0 & 78.6 \\
\midrule
\multirow{3}{*}{\vneipred} & 3 & 46.4 & 78.7 \\
& 5 & \textbf{47.0} & \textbf{78.7} \\
& 10 & 46.2 & 78.6 \\
\bottomrule
\end{tabular}
\vspace{-5pt} 
\caption{\small \textbf{Effects of the number of neighbors on action predictor.} Similar to~\autoref{tab:num_nei_rank}, the size of five is most appropriate for the action prediction. We use \vneitextvisualrank for the ranker.}
\label{tab:num_nei_pred}
\vspace{-15pt}
\end{table}

\mypar{Effects of the number of visual neighbors}
We ablate the impact of varying sizes of visual neighbors, starting with ~\autoref{tab:num_nei_rank}, which shows its effect on the ranker while maintaining the same action predictor (\basepred). We observe a linear correlation between the size of visual neighbors and their ranking/action prediction performance. For instance, 
increasing the size of neighbors up to five shows consistent improvements (\eg, 89.1\%$\rightarrow$90.1\% on Recall@50 and 75.1\%$\rightarrow$78.6\% on Op.~F1). However, considering too many neighbors (\eg, the size of ten) hurts the performance. For example, increasing the size from five to ten decreases the element accuracy from 46.0\% to 45.2\%. We also see a similar pattern when ablating the effect of the visual neighbor size on the action predictor (\autoref{tab:num_nei_pred}). Concretely, while keeping the same ranker (\vneitextvisualrank), the action performance increases up to the size of five (\eg, 46.0\%$\rightarrow$47.0\% on Ele.~Acc) but decreases when the size becomes ten (\eg, 46.2\% on Ele.~Acc). Overall, this suggests that choosing an appropriate number of neighbors is necessary for both element ranking and action prediction.

\section{Conclusion}
\label{sec:conclusion}

We introduce \ourapp to effectively represent HTML elements for web navigation. \ourapp contextualizes each element with its visual neighbor elements, leveraging both textual and visual features. \ourapp consistently improves real-world web navigation in the Mind2Web benchmark, supported by comprehensive analyses.

\section*{Acknowledgments}
This research is supported in part by grants from the National
Science Foundation (IIS-2107077, OAC-2112606) and ARL W911NF2220144. We are thankful for the generous support of the computational resources by the Ohio Supercomputer Center.

{
    \small
    \bibliographystyle{ieeenat_fullname}
    \bibliography{main}
}
\appendix
\section*{Appendices}
In this supplementary material, we provide details omitted in the main text.
\begin{itemize} 
    [itemsep=5pt,topsep=1.5pt]
    \item \autoref{apdx:model_imp_det}: Model implementation \& training details 
    (cf.~\S\ref{ssec:multi_view_ranker}, ~\S\ref{ssec:action_predictor}, and~\S\ref{sec:exp} of the main text).
    \item \autoref{apdx:data_det}: Dataset details (cf.~\S\ref{sec:exp} of the main text).
    \item \autoref{apdx:add_exp}: Additional experiments (cf.~\S\ref{ssec:exp_anal} of the main text).
\end{itemize}

\section{Model implementation \& training details}
\label{apdx:model_imp_det}
As mentioned in~\S\ref{sec:intro} of the main text, we implement \ourapp on top of MindAct algorithm~\cite{mind2web}. We exactly follow its  implementation\footnote{\url{https://github.com/OSU-NLP-Group/Mind2Web}} but provide the details for reference.

\subsection{\ourappbf-enhanced element ranker}
\label{apdx:multi_view_ranker}
MindAct utilizes a small ranking LM to measure the importance of each element $e_{t}$ for action prediction. Concretely, at each time step $t$, the ranking LM takes the element's HTML text tokens $h_{e_{t}}$, the task description $q$, and the previous actions $\{a_1, a_2, \cdots, a_{t-1}\}$ as input and outputs its importance,
\begin{align}
    s_{e_{t}} = f(q, h_{e_{t}}, \{a_1, a_2, \cdots, a_{t-1}\})
\end{align}

\ourapp aims to expand this ranking LM to integrate (i) each element's visual features and textual features and (ii) both the candidate element and its neighbor elements. (See \autoref{fig:ranker} of the main text for an illustration.) 

\mypar{Integrating visual and textual features}
We first extract each element's visual features from the Pix2Struct Vision Transformer (ViT)~\cite{pix2struct}, pre-trained on webpage screenshots. Concretely, Pix2Struct learns rich representations of webpages by asking to predict an HTML-based parse from a masked screenshot. We input the whole screenshot $I_{t}$ to Pix2Struct\textsubscript{base} and apply RoIAlign~\cite{he2017mask} on its output embeddings to obtain the element's visual features $v_{e_{t}}$ based on its bounding box. On the HTML document side, we extract the element's HTML text $h_{e_{t}}$, using the triplet of its ID, HTML text, and bounding box provided in the HTML document.

\mypar{Intergrating visual neighbor elements}
Based on our key insight on webpages---web developers tend to arrange semantically relevant and task-related elements in proximity to each other on the screenshot to enhance user experiences---we contextualize each element $e_{t}$ with its ``visual'' neighboring elements $M_{e_{t}}$. We measure the center points of all elements in the screenshot using their bounding boxes and calculate their pairwise Euclidean distances\footnote{\url{https://scikit-learn.org}}. For each \emph{candidate} element to be ranked by MindAct, we search for the closest $M$ elements to form its context jointly.

\mypar{Aligning visual and textual embedding spaces}
After obtaining each element's visual features $v_{e_{t}}$ and textual features $h_{e_{t}}$, we align them in the same embedding space. Following the recent practice of vision-and-language models (\eg, BLIP-2~\cite{BLIP-2}, LLaVA-1.5~\cite{LLaVA-1.5}), we apply two linear projection layers $W$ to map visual features into the textual embedding space.
We then introduce a learnable positional embedding to (i) pair each projected visual feature $u_{e_{t}}$ with its associated text tokens $h_{e_{t}}$ and (ii) encode the relative distance between the candidate element $e_{t}$ and its neighboring elements $M_{e_{t}}$. Concretely, we add the same positional embedding $p_{e_{t}}$ to the candidate element's (projected) visual feature $u_{e_{t}}$ and textual feature $h_{e_{t}}$. Besides, we sort the neighbors $M_{e_{t}}$ based on their spatial distances from the candidate element $e_{t}$. We then encode the relative positional embedding $p_{m_{e_{t}}^{k}}$ (based on the spatial distance from the candidate) to each neighbor element's visual features $u_{m_{e_{t}}^{k}}$ and corresponding text tokens $h_{m_{e_{t}}^{k}}$. 
We denote the set of the neighbors' visual features by $U_{M_{e_{t}}}$. Similarly, $H_{M_{e_{t}}}$ and $P_{M_{e_{t}}}$ represent the set of their textual features and that of their positional embeddings, respectively.
These positionally encoded visual and textual token embeddings (of the candidate and the neighbor elements) are passed into the ranking LM $f$; the visual features are prepended to the textual embeddings, serving as soft visual prompts, 

\begin{equation}
\begin{aligned}
    & s_{e_{t}} = f(q, R_{e_{t}}, \{a_1, a_2, \cdots, a_{t-1}\}), \\
    & R_{e_{t}} = [u_{e_{t}}+p_{e_{t}};
    U_{M_{e_{t}}}+P_{M_{e_{t}}};
    h_{e_{t}}+p_{e_{t}};
    H_{M_{e_{t}}}+P_{M_{e_{t}}}]
\end{aligned}
\end{equation}

\begin{table*}[t]
\centering
\small
\tabcolsep 8.5pt
\renewcommand\arraystretch{1.0}
\begin{tabular}{cccccccc}
\toprule
\multirow{2}{*}{Dataset} & 
\multirow{2}{*}{\# Domains} &
\multirow{2}{*}{\# Websites} &
\multirow{2}{*}{Website} & 
\multirow{2}{*}{\# Tasks} & 
\multirow{2}{*}{Avg \#} &
\multicolumn{2}{c}{Avg \# HTML} \\
\cmidrule(r){7-8}
& & & Type & & Actions & Elements & Tokens \\
\midrule
MiniWoB++~\cite{Humphreys2022ADA} & - & 100 & Simplified & 100 & 3.6 & 28 & 500 \\
Mind2Web~\cite{mind2web} & 31 & 137 & Real-world & 2,350 & 7.3 & 1,135 & 44,402 \\
\bottomrule
\end{tabular}
\vspace{-5pt}
\captionsetup{width=0.9\textwidth} 
\caption{\small \textbf{Detailed Statistics of Mind2Web~\cite{mind2web}.} Min2Web is the first real-world web navigation benchmark, collecting over 100 real-world websites across various domains. Unlike previous benchmarks~\cite{Humphreys2022ADA,Yao2022WebShopTS}, Mind2Web provides an extensive amount of real-world webpage content, including over 1K/44K HTML elements/tokens on average.}
\label{apdx_tab:mind2web_stat}
\vspace{-10pt}
\end{table*}

\mypar{Training Details}
In training, we only learn the projection layer $W$, the positional embeddings $P$, and the ranking LM $f$ while keeping the ViT frozen. For the ranking LM, we use DeBERTa\textsubscript{base}~\cite{he2020deberta}, a small encoder-only LM. We exactly follow the configuration of MindAct. Specifically, we train the LM (together with a linear classifier)  with a batch size of 32 and a learning rate of 3e-5 for 5 epochs. The LM outputs the element's importance score through a sigmoid activation
function. The score is optimized with a binary cross-entropy loss, where 
the ground-truth element serves as a positive example, and elements randomly sampled from the webpage are considered negative examples. The LM is trained on a single Nvidia A6000 48GB GPU.
During inference, we score all candidate elements in the webpage and select top-$K$ elements for the action predictor.

\subsection{\ourappbf-enhanced action predictor}
Due to the high computational cost of directly passing an entire HTML document into LLMs, MindAct~\cite{mind2web} restricts its input to only the top-$K$ candidate elements selected from the ranking LM. Concretely, MindAct combines the selected elements into an HTML snippet $H_{t}$ and feeds it into an LLM $g$, along with the task description $q$ (``Find one-way flights from New York to Toronto.'') and the previous actions $\{a_1, a_2, \cdots, a_{t-1}\}$ (``Type New York in the From box''). At each time step $t$, the objective is to predict an action $a_{t}$, composing of the target element $e_{t}$ (\eg, ``[textbox] To'') and its associated operation $o_{t}$ (\eg, ``Type Toronto''),
\begin{equation}
\begin{aligned}
    &a_{t} = g(q, H_t, \{a_1, a_2, \cdots, a_{t-1}\}), \\
    &a_t: \{e_{t}, o_{t}\}
\end{aligned}
\end{equation}
We note that MindAct converts the target element prediction problem into multiple-choice question-answering. Instead of directly generating the target element, they split top-$K$ candidates into multiple clusters of five element options (including the ``None'' option) and ask the LLM to pick one element from each cluster. If more than one element is selected, they form a new group with the chosen ones and iterate this process until a single element is selected.

The action predictor of \ourapp takes the same input as MindAct, except for appending each candidate element with its neighboring elements. We generate an HTML snippet $S_t$ based on the top-$K$ candidate elements and their adjacent elements, and input the snippet (with the task description and the previous actions) to the LLM $g$ and predict the action $a_t$,
\begin{equation}
\begin{aligned}
    &a_{t} = g(q, S_t, \{a_1, a_2, \cdots, a_{t-1}\})
\end{aligned}
\end{equation}

\mypar{Training Details}
We again adopt the configuration from MindAct. We train Flan-T5\textsubscript{base}~\cite{flan-t5}, an instruction fine-tuned encoder-decoder LLM, with a batch size of 32 and a learning rate of 5e-5 for 5 epochs. We optimize its parameters with the language modeling loss on a single Nvidia A6000 48GB GPU.

\section{Dataset Details}
\label{apdx:data_det}
Mind2Web~\cite{mind2web} recently proposed the first real-world web navigation benchmark, consisting of over 2,000 open-ended tasks from more than 100 real-world websites. They collect the websites across 31 diverse domains, including travel, shopping, entertainment, public service, etc. Unlike other existing benchmarks~\cite{Humphreys2022ADA,Yao2022WebShopTS} limited to simulated environments, Mind2Web instead focuses on real-world environments (\autoref{apdx_tab:mind2web_stat}). For instance, Mind2Web provides real-world websites with rich content, including thousands of HTML elements, tens of thousands of HTML tokens, and 7.3 web-related actions per task on average.

\mypar{Data Collection}
Given a real-world website (\eg, an airline website), Mind2Web first asks annotators to write open-ended realistic tasks (\eg, ``Find one-way flights from New York to Toronto.'') relevant to the website. The workers are then required to complete the defined task with a sequence of actions. Specifically, each action is composed of element selection and operation selection. The annotators should first find an element (\eg, ``[textbox] From'') relevant to the task on the webpage and perform an operation (\eg, ``Type New York'') on the element.

\mypar{Dataset Split}
The Mind2Web dataset provides a training split with 1,009 real-world tasks collected from 73 websites. Each task consists of a sequence of action samples. In total, there exist 7,775 samples in the training split. Mind2Web evaluates a web agent on three different test splits.
\textbf{Test\textsubscript{Cross-Domain}} measures the agent's generalizability to a new domain where it has not seen any websites or tasks associated with that domain during training. The split contains 912 tasks with 5,911 samples from 73 real-world websites.
In \textbf{Test\textsubscript{Cross-Website}}, while the agent is not exposed to test websites, it is trained on websites from the same domain and potentially with similar tasks. This configuration enables us to evaluate the agent's capacity to adapt to entirely new websites within familiar domains and tasks. This split consists of 177 tasks, along with 1,373 samples obtained from 10 websites.
\textbf{Cross-Task} is a conventional test split, which is the random 20\% of the dataset. The split has 252 tasks with 2,094 samples from 69 websites.

\mypar{Task Details}
The Mind2Web task consists of a sequence of actions, each comprising a pair of an actionable HTML element (\eg, ``[textbox] To'') and an operation (\eg, ``Type Toronto''). Mind2Web provides three common operations: Click, Type, and Select. For Type and Select operations, an additional argument (\eg, ``Toronto'') is required.

\section{Additional Experiments}
\label{apdx:add_exp}

\mypar{More powerful action predictor}
We scale up the predictor from Flan-T5\textsubscript{base} to Flan-T5\textsubscript{large} to check whether our visual neighbors are still beneficial with the larger model. As shown in~\autoref{larger_predictor}, \ourapp still achieves notable gains, suggesting the complementary capabilities of LLMs and our visual neighbors.

\begin{table}[t]
\centering
\small
\tabcolsep 2.5pt
\renewcommand\arraystretch{1.0}
\begin{tabular}{llccc}
\toprule
\multirow{2}{*}{Ranker} &
\multirow{2}{*}{Action} &
\multicolumn{3}{c}{Cross-Task} \\
\cmidrule(r){3-5}
& Predictor & Ele. Acc & Op. F1 & Step SR \\
\midrule
\multirow{2}{*}{\baserank} & \basepredlarge & 51.4 & 75.6 & 48.7 \\
& \vneipredlarge & 54.2 & 79.5 & 50.9 \\
\bottomrule
\end{tabular}
\vspace{-5pt} 
\caption{\small \textbf{\ourappbf with a larger predictor.}
We increase the size of the predictor from Flan-T5\textsubscript{base} to Flan-T5\textsubscript{large}. Even with the larger predictor, \ourapp notably outperforms the baseline, showing the complementarity of \ourapp and LLMs.}
\label{larger_predictor}
\vspace{-10pt}
\end{table}

\mypar{Neighbors from an HTML tree}
An HTML document can be represented as a DOM tree, a hierarchical tree of HTML objects (\eg, Element: <head>).
Thus, we can also extract each element's neighbors from the HTML tree. We compare the tree-based neighbors with 
our neighbors obtained from the screenshot (\autoref{apdx_tab:add_whole_data}). Our visual neighbors (\vneipred) significantly outperform those defined by the HTML tree (\htmltreeneipred), suggesting that visual-spatial context is more beneficial.

\mypar{Ranker with whole visual tokens} In~\S\ref{ssec:exp_anal} of the main text, we show that \ourapp (\ie, the use of visual neighbors) is more effective than the use of the entire image for web navigation (\eg, \vneipred vs.~\wholeimagepred, \vneitextvisualrank vs.~\wholeimagerank). 
To further substantiate the efficacy of \ourapp over using the whole image, we conduct additional experiments (\autoref{apdx_tab:add_whole_data}). Specifically, we train a ranker (\wholevistokrank) using \emph{all visual tokens} extracted from the whole image based on the Pix2Struct ViT~\cite{pix2struct}. Like the previous results in the main text, \wholevistokrank outperforms the baseline (\eg, 44.1\% vs.~42.0\%), suggesting the benefit of utilizing the entire image. However, \wholevistokrank falls short of \vneitextvisualrank (46.0\%), which uses significantly fewer inputs (\ie, only neighboring elements). This again supports the advantages of \ourapp over the whole image regarding computational efficiency and performance.

\begin{table}[t]
\centering
\small
\tabcolsep 4pt
\renewcommand\arraystretch{1.0}
\begin{tabular}{llc}
\toprule
\multirow{2}{*}{Ranker} &
Action &
Cross-Task \\
& Predictor & Ele. Acc \\
\midrule
\multirow{3}{*}{\baserank} & \basepred & 42.0 \\
& \wholeimagepred & 43.6 \\
& \htmltreeneipred & 43.8 \\
& \vneipred & 44.4 \\
\midrule
\wholeimagerank & \multirow{3}{*}{\basepred} & 43.9 \\
\wholevistokrank & & 44.1 \\
\vneitextrank & & 44.6 \\
\vneitextvisualrank & & \textbf{46.0} \\
\midrule
\quad - & \wholehtmlpred & 38.6 \\
\bottomrule
\end{tabular}
\vspace{-5pt} 
\caption{\small \textbf{Additional results for \autoref{tab:whole_data} in the main text.} Our neighbors defined by a screenshot (\vneipred) notably outperform the neighbors defined by an HTML tree (\htmltreeneipred). Moreover, \vneitextvisualrank is significantly better than \wholevistokrank, which uses all visual tokens of the entire image. This again highlights the benefit of \ourapp in both computational efficiency and performance.}
\label{apdx_tab:add_whole_data}
\vspace{-10pt}
\end{table}

\mypar{Type of pre-trained visual features}
\autoref{apdx_tab:vis_feat_type} summarizes the importance of the type of pre-trained visual features on web navigation. As discussed in~\S\ref{ssec:context_enh} of the main text, to train the ranker, we extract the element's visual features using Pix2Struct~\cite{pix2struct}'s VIT, pre-trained on webpage screenshots. We investigate if these pre-trained ``screenshot'' visual features (\vneitextvisualrankweb) indeed contain meaningful HTML context for downstream web navigation tasks. Concretely, we compare them with features extracted from ViT pre-trained on COCO~\cite{coco}, an object recognition benchmark containing common objects in ``natural images''. We denote a ranker using the COCO visual features by \vneitextvisualrankcoco. We first observe that \vneitextvisualrankcoco outperforms \vneitextrank that only leverages elements' HTML text features to train the ranker (\eg, 45.2\% vs.~44.6\% on Ele.~Acc). This implies that even if visual features are from a different domain (\ie, natural images), incorporating them is still helpful in web navigation tasks.
However, compared to \vneitextvisualrankweb, which uses both HTML visual and textual features, \vneitextvisualrankcoco performs less (\eg, 46.0\% vs.~45.2\% on Ele.~Acc). This highlights that the pre-trained ``screenshot'' visual features indeed contain HTML-related context, which benefits more in completing the downstream web navigation tasks.
\begin{table}[t]
\centering
\small
\renewcommand\arraystretch{1.0}
\begin{tabular}{lccc}
\toprule
\multirow{2}{*}{Ranker} &
\multicolumn{3}{c}{Cross-Task} \\
\cmidrule(r){2-4}
& Ele. Acc & Op. F1 & Step SR \\
\midrule
\vneitextrank & 44.6 & 75.7 & 43.2 \\
\vneitextvisualrankcoco & 45.2 & 76.3 & 43.4 \\
\vneitextvisualrankweb & \textbf{46.0} & \textbf{78.6} & \textbf{44.8} \\
\bottomrule
\end{tabular}
\vspace{-5pt} 
\caption{\small \textbf{Effects of different types of pre-trained visual features.} The pre-trained screenshot visual features~\cite{pix2struct} are more beneficial on the downstream web navigation than those extracted from ViT pre-trained on natural images of COCO~\cite{coco}.}
\label{apdx_tab:vis_feat_type}
\vspace{-10pt}
\end{table}

\mypar{Existing/Concurrent Works}
A number of previous studies~\cite{Humphreys2022ADA,Yao2022WebShopTS,pixelhelp,meta-gui,burns2022dataset,liu2018reinforcement,jia2019dom,ASH,pix2act} have explored web navigation but mainly worked on \emph{simplified} websites~\cite{Humphreys2022ADA,Yao2022WebShopTS}, which deviate from the focus of our study. Our attention is instead directed towards \emph{real-world} scenarios involving various real-world websites with extensive raw HTML documents (\eg, Mind2Web).
We have identified a few \emph{concurrent} works~\cite{WebGUM,Gur2023ARW,zheng2024gpt,he2024webvoyager,hong2023cogagent,cheng2024seeclick} exploring Mind2Web, but they mostly focus on (i) large-scale pre-training, requiring substantial amounts of pre-training HTML data, or (ii) evaluating the potential of recent vision-and-language models (\eg, GPT4-V~\cite{gpt4}) as a web agent. As their codes or pre-training datasets have not been released yet, replicating their work would be prohibitively costly. We thus do not consider them in our studies.

\end{document}